\documentclass[letterpaper, 10 pt, conference]{ieeeconf}

\IEEEoverridecommandlockouts

\usepackage{amsmath,amssymb,amsfonts}
\usepackage{algorithmic}
\usepackage{graphicx}
\usepackage{textcomp}
\usepackage{xcolor}
\usepackage{subcaption}
\usepackage[allowmove]{url}

\usepackage{xcolor}
\usepackage{changes}
\DeclareMathOperator*{\argmax}{argmax}

\usepackage[capitalise]{cleveref}

\title{\LARGE \bf
Self-supervised quality evaluation of grid maps without ground truth}

\author{Unal Artan, Manuel Castellano-Quero, Johannes A. Stork and Martin Magnusson%
\thanks{The authors are with the Center for Advanced Autonomous Sensor Systemes (AASS), \"{O}rebro University, Fakultetsgatan 1, Sweden. 
Corresponding author {\tt\small unal.artan@oru.se}.
This work has received funding from the Swedish Knowledge Foundation (KKS) project ”NiCE” under agreement number 20200247 01 H, 
the Knut and Alice Wallenberg Foundation through Wallenberg AI, Autonomous Systems and Software Program (WASP),
the Swedish Knowledge Foundation in the TeamRob Synergy under Project 20210016,
and the European Union’s Horizon 2020 research and innovation programme under grant agreement No 101017274 (DARKO).}%
}

\begin{document}

\maketitle
\thispagestyle{empty}
\pagestyle{empty}

\begin{abstract}
Maps are widely used in the robotics community for localization and path planning and the ability to identify incorrect or inaccurate regions of a map is an area of active research interest. For large environments, manual edits of the map can be very time consuming and a tool which can assist a map editor is needed. 
With this paper, we contribute to robot \emph{introspection} through self-assessment of map quality, using only a given map as input to be independent of the underlying mapping algorithm.
Current methods for map quality assessment rely on ground-truth or labelled information for training classifiers, both of which can be difficult to obtain. 
In this paper, we propose a \emph{self-supervised learning approach} to map quality assessment of 2D occupancy grid maps which identifies regions of a map that require further investigation. We leverage an autoencoder for patches extracted from the map to capture the dominant structure of the map. By evaluating the reconstruction loss of a patch reconstructed with the learned model, we can evaluate whether it contains an unusual structure, which indicates the need of further investigation. We evaluate our method on the outcome of state-of-the-art grid mapping methods using benchmark datasets as well as maps from warehouse environments, highlighting the application to more complex environments. We show that the method identifies regions with sparse data, clutter, and misregistrations.

\end{abstract}

\section{Introduction}
Robot maps need to be accurate, especially in industrial environments, where safety and efficiency of operation is of the utmost importance. Therefore, any errors related to mapping should be detected and fixed, since the use of incorrect or inaccurate maps in applications such as navigation or path planning could severely compromise the robotic operation. Cleaning and repairing a map can be a time-consuming task, especially if performed manually, since each environment may pose different challenges, and the ability and experience of the map editor would be key in that case to ensure all mapping errors are corrected adequately. For this reason, a more automated approach would be desirable, however, such a tool is still not readily available.

One could make the case that map quality assessment is not needed given recent advances in localization algorithms \cite{Zhao-RSS-22,AnSu2022,KISS_ICP,MAD_ICP} and improvements to sensing and infrastructure \cite{adolfsson-2023-tbv,kubelka-2024-radar,zhang20234dradarslam}. However, while they may reduce the errors mentioned previously, the advances have yet to remove these errors completely \cite{Zhao-RSS-22,KISS_ICP,MAD_ICP}, hence the present need for a map quality assessment tool. This is evident by assessing leader boards to autonomous driving challenges where the leaders have non-zero error \cite{KITTI_Dataset,OxfordDataset}. 

\begin{figure}[htbp]
    \centering
    \includegraphics[width=0.485\columnwidth]{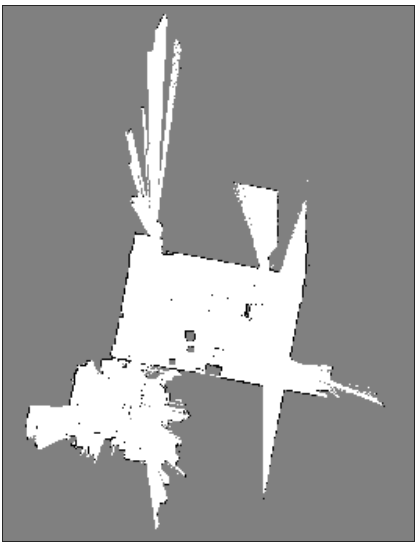}~
    \includegraphics[width=0.485\columnwidth]{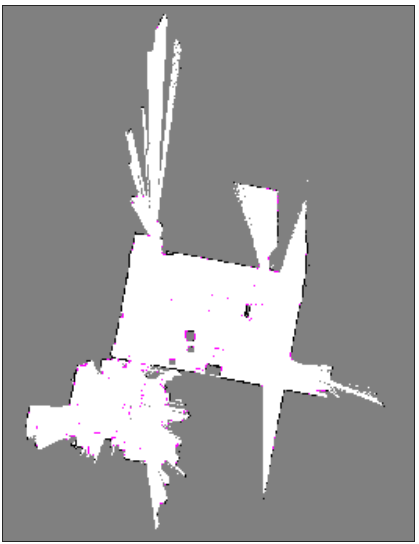}        
    \caption{Example output from a modern grid-map based SLAM algorithm (\url{https://github.com/wangyingyu/Occupancy-SLAM} \cite{Zhao-RSS-22}). While the result is a map that is generally consistent, there are several areas with faults which are accurately indicated by our evaluation method. Map cells identified as suspicious by our method are highlighted in magenta (right).  In particular, the room in the lower part of the map contains misregistrations and is sparsely covered. Some of the stray obstacle cells in the large room also warrant further inspection.}
    \label{fig:occupancy-slam-subfigure-1}
\end{figure}

Errors may manifest in a map in different ways. For instance, local misregistration often appears as doubled or slightly bent walls. Errors due to faulty loop closure may appear as overlapping map segments. More subtly, errors due to sparse coverage may manifest as poorly reconstructed or partially missing obstacles. Other errors would include remnants of moving obstacles and clutter that were not properly cleared from the map (see \cref{fig:occupancy-slam-subfigure-1}).

Performing map quality assessment in an autonomous manner would help to reduce the cognitive load of a user verifying a map during commissioning of a new robot system, and it would also serve to assist the editing process to ensure the final map is as accurate as possible. It could ultimately be used by the robot itself to go back and improve the identified places. In this context, an \emph{introspective} map assessment tool that relies only on the map without the need of additional information, e.g., labels or ground truth reference, would be desirable.

A potential approach to map quality assessment could be to develop an analytical solution that captures the wide range of potential map errors. However, this process would be time consuming and would rely on the ability of the human expert generating the models. The problem has also been formulated using image classification, where map segments or patches are extracted from different maps and labelled either manually \cite{andersone-2020-quality} or with some ground truth (i.e., in simulation) to train and evaluate a convolution neural network. Similar to using an analytical solution, this type of methods relies on the quality of labelling. Additionally, applying the mentioned methods for differing environments such as an indoor office and an underground mine would require two unique sets of analytical methods and two sets of labelled patches, which are both time consuming tasks.

We hypothesize and which we demonstrate that large regions of built maps are \emph{usually} correct and that typically only small regions are inaccurate. With this in mind, we approach map quality assessment as a sort of \emph{self-supervised} anomaly detection problem \cite{hawkins1980identification,pang2021deep}. We propose the use of a data driven approach to learn a model of the expected appearance of the structure in the map, while we consider the rest as \textit{anomalies}. Consequently our approach removes the necessity for labelled training data.

\section{Related Work}

Robotic grid maps are a product of several inputs (such as the robot's pose and local sensing information). Minimizing these input errors is a way to ensure robotic maps are as accurate as possible and to reduce the amount of map errors \cite{Kazerouni2022,nubert2022}. In this sense, Zhao et al. \cite{Zhao-RSS-22} recently proposed an offline, optimization-based approach for SLAM in which both the robot poses and the occupancy map are simultaneously optimized, leading to more accurate trajectory and map estimations.
While reducing errors during the mapping process is important in its own right, methods like these are not helpful for assessing the quality of a given map. 

Early work on robot mapping has been using  entropy as a measure of map goodness \cite{saez-2005-entropy,stachniss2003exploring,stachniss2005information}. 
If online sensor input is available, a change in entropy can also be used to indicate registration errors \cite{adolfsson-2022-coral}.
However, entropy alone is not an effective measure of the quality of a given map, as demonstrated in our experiments in \cref{fig:Map-B-entropy} of \cref{sec:experimental-evaluation}.

Clutter, a form of map error, needs to be removed from a map. For indoor environments that are structured, wall directions can be used to filter clutter and non-structure features (see e.g., \cite{Kucner2021,Luperto1691663}). These works rely on a 2D Discrete Fourier Transform (DFT) of the map to identify the set of wall angles present in the map \cite{Kucner2021}. With this knowledge, map cells are then classified as \textit{structure} or \textit{clutter}. Also, boundary and free-space noise are sources of mapping errors. They are addressed in the work by An and Kim \cite{AnSu2022}, which introduces a novel method for global localization based on the extraction of statistical signatures of geometry and structure in 2D occupancy grid maps. A score based on these features is proposed to measure similarity between pairs of submaps for localization, and it is demonstrated to be robust against both boundary and free-space errors. However, this metric does not provide an assessment of the quality of maps.

Map quality assessment with ground truth maps has been explored \cite{Moravec1993,Wagan2008,Varsadan2009,Schwertfeger2012} and relates map quality to a similarity metric calculated using ground truth map and the generated map.
However, such reference-based quality assessment is mostly useful in a benchmarking or competition setting. 
For evaluating the quality of a map in a real-world deployment, reference-free methods are required.

Thus far, learning-based approaches to map quality assessment have been limited to supervised methods. In Andersone~\cite{andersone-2020-quality}, for instance, labelled map patches are employed to train a neural network binary classifier, which yielded good results in identifying incorrect regions of a grid map. While the aim was to provide a map quality metric for assessing map merging, the binary classifier performed well. A drawback to this approach is the need for labelled map patches, which can be difficult to obtain. Additionally, constraints such as maximum number of unknown cells and minimum number of cells occupied in a patch could lead to significant regions of a map to be not evaluated.

A common theme in the work of Andersone~\cite{andersone-2020-quality} and Kucner et  al.~\cite{Kucner2021} is to classify using a threshold or binary classification. Binary classification can also be thought of as an anomaly detection problem and is the approach we pursue in this paper. Image-based anomaly detection using deep learning methods \cite{Zhou2017,Beggel2020,Shvetsova2021,pang2021deep} has yielded promising results and confirms our direction of treating map quality assessment as an anomaly detection problem.

\section{Proposed Method}

We pose map quality assessment as a deviation-based anomaly detection problem. For this, we consider the grid map on the level of individual patches $\mathbf{P}$ and learn a low-dimensional embedding space that only represents valid patches without noise or distortion. This is an unsupervised learning problem. Mapping patches $\mathbf{P}$ to this space and comparing them to their reconstruction $\mathbf{P}^\prime$ allows us to detect abnormal patches based on the reconstruction error.

\subsection{Training Data}

The training data for learning are patches $\mathbf{P}$ from occupancy grid map(s) \cite{Elfes1989}, which can be uniquely identified by an offset coordinate and a width and a height. Several patches might overlap a cell in the map. The training patches for learning the embedding (i.e., the encoder and decoder mappings) could be obtained from a different map than the map being assessed. However, both maps would need to represent the same general type of environment (otherwise the general characteristics of the maps may not be similar enough). Formulating map quality assessment as an anomaly detection problem relies on the grid map(s) for extracting training patches are mostly correct. However, except for degenerate cases, the majority of patches in a map can be expected to be correct.

\subsection{Deviation-based Anomaly Detection with Autoencoder}
\label{sec:cvae}

We use an image-based convolutional variational autoencoder (VAE) \cite{kingma2014auto} where the inputs are map patches, $\mathbf{P}$, extracted from the occupancy grid map. The objective of the autoencoder is to minimize the reconstruction error and match the latent space prior $p(\mathbf{z})$ by jointly optimizing the generative parameters $\theta$ and embedding parameters $\phi$. The encoder network maps from patches to Gaussians in the latent space $q_\phi(\mathbf{z} \mid \mathbf{P}) = \mathcal{N}(\mu_\phi(\mathbf{P}), S_\phi(\mathbf{P}))$ and the decoder network maps from latent space positions to patches $f_\theta(\mathbf{z}) = \mathbf{P}^\prime$. The evidence lower bound objective (ELBO) can be stated as
\begin{equation}
\mathcal{L}_{\theta, \phi}(\mathbf{P})
=
\mathbb{E}_{\mathbf{z} \sim q_\phi(. \mid \mathbf{P})}
[
-\mathcal{L}(\mathbf{P}, \mathbf{P}^\prime)
]
-
D_\mathit{KL}(q_\phi (. \mid \mathbf{P})\parallel p(.))
,
\end{equation}
where we assume the prior $p(\mathbf{z})$ as Gaussian and $D_{\mathit{KL}}$ the Kullback–Leibler divergence \cite{Kullback1951}.

The reconstruction loss for patches $\mathcal{L}_{\theta, \phi}(\mathbf{P})$ is implemented as
\begin{multline}
   \mathcal{L}_{\theta, \phi}(\mathbf{P}) =   \argmax_{\theta,\phi} \\ \left(
\mathbb{E}_{\mathbf{z} \sim q_\phi(. \mid \mathbf{P})} 
\left ( \frac{\|
\mathbf{P} -  \mathbf{P}^\prime
\|^{2}}{2c} \right )
- \
D_\mathit{KL}(q_\phi (. \mid \mathbf{P})\parallel p(.)) \right ).
\end{multline}
The parameters are learned end-to-end using the re-parameterization trick for backpropagation \cite{kingma2014auto}.

Given a sufficiently large network capacity and time the reconstruction loss would be reduced to zero. However, this isn't the case... 

\subsection{Grid Cell Classification}
\label{sec:cell-classification}
Our approach exploits the fact that the majority of cells in the training data are accurate and are selected so as to cover the whole environment. The autoencoder model, trained with these patches, has learned the dominant structure of cells within the data. Identifying the inaccurate cells using the reconstruction loss is the next step. The autoencoder is trained on patches, but these are solely a compilation of map cells, therefore, we do not consider the total reconstruction error of a particular patch, which would be similar to using patch Entropy, but we consider each cell in it for classification.

A map cell $\mathrm{m}_{k}$ is classified as either \textit{good} (plausible) or \textit{bad} (implausible) by evaluating the mean absolute reconstruction loss $e_{k}$ estimated using all of the patches $\mathbf{P}$ and reconstructed patches $\mathbf{P^{\prime}}$ that overlap with the map cell, which we denote as $\mathbf{\tilde{P}}_k$. That is, for a set of patches $\mathbf{\tilde{P}}_{k}$ that overlap with the map cell $m_{k}$ the mean absolute reconstruction loss for the cell is
\begin{equation}\label{eq:loss-cell}
    e_{k} = \frac{1}{N_{\tilde{P}_{k}}} \sum_{k=1}^{k=N_{\tilde{P}_{k}}} \lvert \mathbf{P} - \mathbf{P^{\prime}} \rvert,
\end{equation}
where $N_{\tilde{P}_{k}}$ is the number of patches that overlap map cell $\mathrm{m}_{k}$.

A map cell is then classified as \textit{good} when $e_{k} <= e_{\tau}$ and \textit{bad} when $e_{k} > e_{\tau}$. Given the earlier assumption that the majority of the map cells are accurate, we calculate numerically the threshold $e_{\tau}$ that corresponds to 5~\% of $\mathbf{e}$ with map cell reconstruction errors larger than $e_{\tau}$ with $e_{\tau}$ a tunable hyperparameter.

\section{Experimental Evaluation}
\label{sec:experimental-evaluation}
We demonstrate the effectiveness of the proposed method by first highlighting that Entropy is not an effective measure of the quality of a given map. Secondly, we comparing classification to that of Andersone~\cite{andersone-2020-quality} that used a neural network classification trained using manually labelled map patches. Thirdly, we apply our approach to a map generated using a state-of-the-art grid mapping technique and to maps of real-world robot deployments in warehouse environments. We assess whether a model trained only on one map can be used to evaluate maps of different environments.

\subsection{Data Conditioning and Training Parameters}

In occupancy grid maps, cell occupancy, $O$, is in the range $O = [0,1]$ with the extremes representing \textit{free} and \textit{occupied} or vice-versa. Map cells are commonly initialized with an occupancy value of $O=0.5$ and discretized to three classes. Our method handles both cases of cell occupancy. 

Patches are extracted from the map using a stride length of two cells. If the patch satisfies the requirements that at most 40~\% of patch cells are \textit{unknown} and at least two of the patch cells are \textit{occupied} then it is included in the training set. The significant cell requirement is the same as those of Andersone~\cite{andersone-2020-quality} where we use a less restrictive occupied cell rate than Andersone's. 

An ablation study is performed, using the SDR site B map shown in \cref{fig:baseline-2-occ-grid} and results given in \cref{tbl:ablation-study}, exploring the network design to the training time, training loss and the cut-off threshold $e_{\tau}$. The study held constant the patch size at $32 \times 32$, batch size of 128 and learning rate of 0.001.
\begin{table}[htbp]
\centering
\begin{tabular}{ |p{4cm}|p{1.25cm}|p{0.85cm}|p{0.75cm}|  }
    \hline
    Parameter Set & Training Time [mins] & Training Loss & $e_{\tau} @$ 95\%\\
    \hline
    \{128,3,2\},64,20   & 4.5   &   33    &  0.404 \\
    \hline
    \{128,5,4\},64,20   & 6.8   &   32     &  0.396 \\
    \hline
    \{32,3,2\},\{64,3,2\},16,20   & 6.2   &   31     &  0.410 \\
    \hline
    \{64,3,2\},\{128,3,2\},16,20   & 7   &   31     &  0.406 \\
    \hline
    \{64,3,2\},\{128,3,2\},32,20   & 6.7   &   32     &  0.397 \\
    \hline
    \{64,3,2\},\{128,3,2\},64,20   & 7.3   &   33     &  0.404 \\
    \hline
    \{64,3,2\},\{128,3,2\},64,200   & 71.7   &   29     &  0.340 \\
    \hline
    \{32,3,2\},\{64,3,2\},\{128,3,2\},32,20   & 8.8   &   30     &  0.372 \\
    \hline
    \{32,3,2\},\{64,3,2\},\{128,3,2\},64,100   & 42.5   &   27     &  0.301 \\
    \hline
\end{tabular}
\caption{Influence of training parameters and training loss, training time and threshold $e_{\tau}$ at 95\%. Values in the column Parameter Set corresponds to: \{Layer $i$ Filter size, Kernel Size, Stride Length\}, \# of Latent Parameters, \# of Epochs and where $i$ increments to the number of layers.}
\label{tbl:ablation-study}
\end{table}

From the ablation study, our method is rather insensitive to the parameter set. Training loss is contained in a narrow range, as is the $e_{\tau}$ threshold. We chose a final encoder network to first downsample the patch with two blocks of 2-D convolutions and ReLU layers with 32 and 64 filters, respectively. A kernel size of $3 \times 3$ and stride length 2 is used for each convolution layer. The latent parameter size is 16. The decoder network has two blocks of transposed convolution and ReLU layers each with kernel size $3 \times 3$ and stride length 2 with 64 and 32 filters, respectively. The Adam optimizer \cite{Kingma2015} with learning rate of 0.001, patch batch size of 128 and 20 epochs is used for training. 

\subsection{Comparison to Entropy Metric}
\begin{figure}[htbp]
    \centering
        \centering    
        \includegraphics[width=0.75\columnwidth]{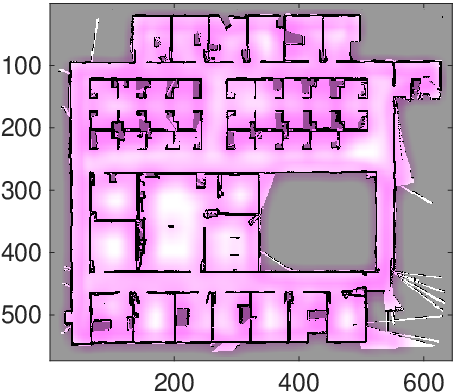}
        \label{fig:Map-B-entropy}
    \caption{SDR site B grid map (Radish: Robotics
Research Datasets (\url{https://dspace.mit.edu/handle/1721.1/62245}) evaluated using patch Entropy.
 While mapping errors often increase entropy, regions with high entropy (complex geometry) do not generally correspond to map errors.}
    \label{fig:baseline-2-occ-grid}
\end{figure}

Patches extracted from the map given in \cref{fig:baseline-2-occ-grid} are analyzed using Entropy to evaluate the possibility of a statistical metric, which is less complex compared to our proposed  autoencoder, could perform map quality assessment.
Several regions that are correct but contain complex geometry are marked as \textit{bad} when measuring the entropy alone, see e.g. the structure inside the array of rooms in the upper part of \cref{fig:baseline-2-occ-grid}.

\begin{figure}[htpb!]
    \centering
    \includegraphics[width=0.4\columnwidth]{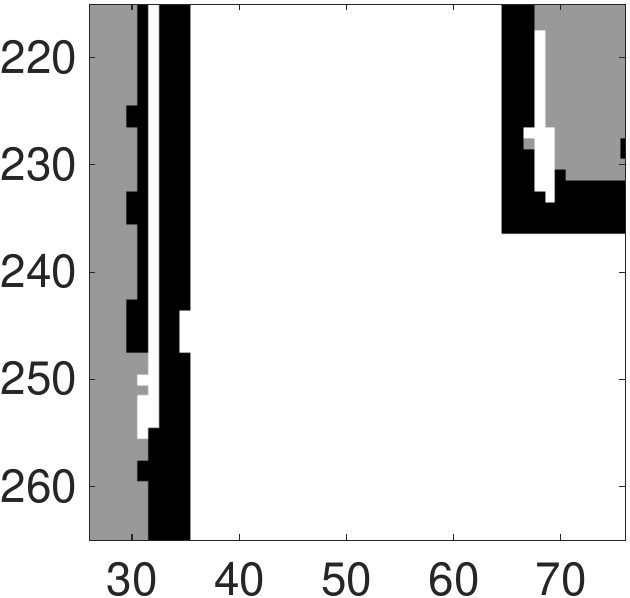}~
    \includegraphics[width=0.4\columnwidth]{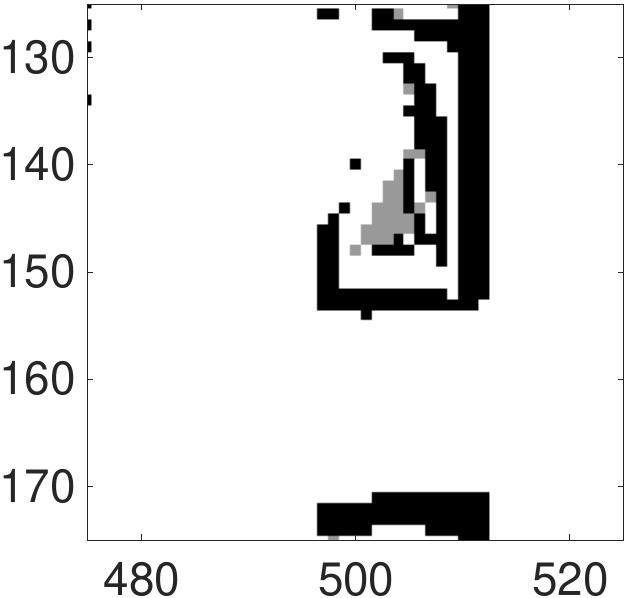} 
    \includegraphics[width=0.4\columnwidth]{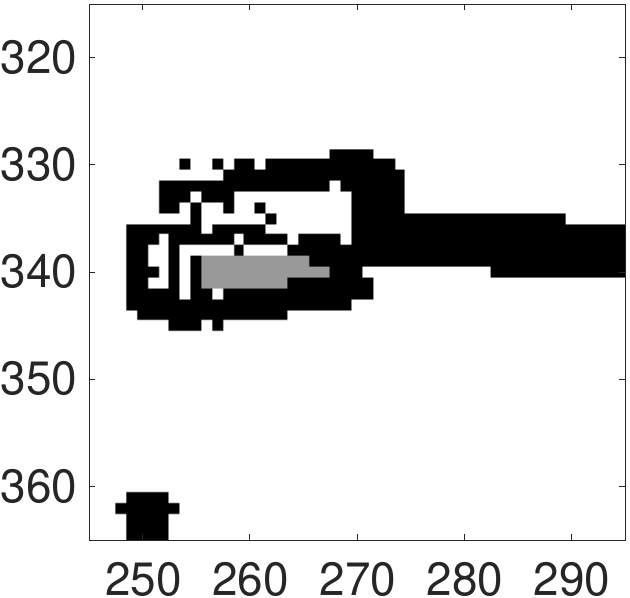}~
    \includegraphics[width=0.4\columnwidth]{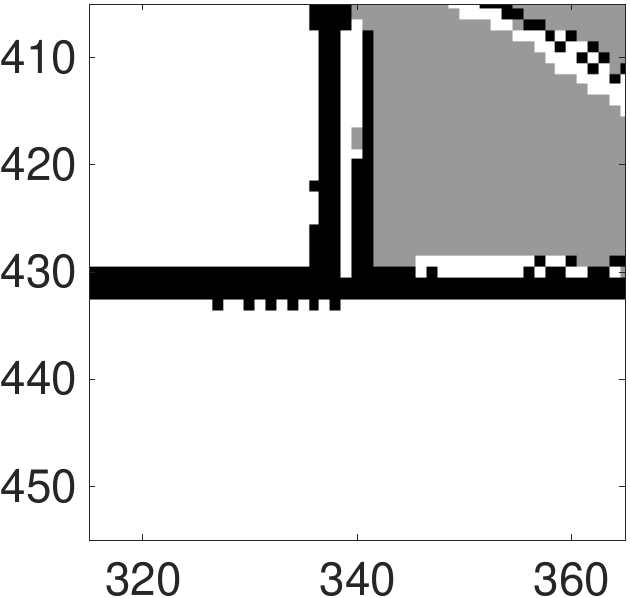}        
    \caption{Zoomed section of the SDR site B map highlighting inaccuracies in the map.}
    \label{fig:Map-B-zoomed-regions}
\end{figure}

The grid map in \cref{fig:baseline-2-occ-grid} is rated as high quality by Andersone~\cite{andersone-2020-quality} and at a large zoom appears to be an accurate assessment. However, on closer inspection, inaccurate map cells regions such as partial observation, double walls, inconsistent wall thickness and free cells between unknown and occupied cells are observed in \cref{fig:Map-B-zoomed-regions} 
This highlights the difficult and time consuming task a map editor encounters when evaluating maps and the need for an automated solution.

\subsection{Comparison to Supervised Learning}
\label{sec:baseline-results}
\begin{figure}%
    \centering
    \begin{subfigure}{\columnwidth}
        \centering
        \includegraphics[width=\columnwidth]{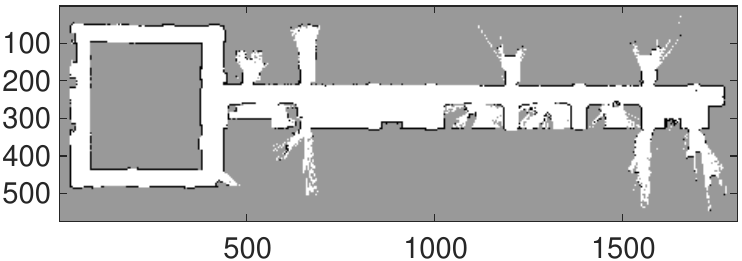}
        \caption{Map A: grid map with partially observed regions and minimal cell inaccuracies.}
        \label{fig:baseline-1-occ-grid}
    \end{subfigure}
    \begin{subfigure}{\columnwidth}
        \centering
        \includegraphics[width=\columnwidth]{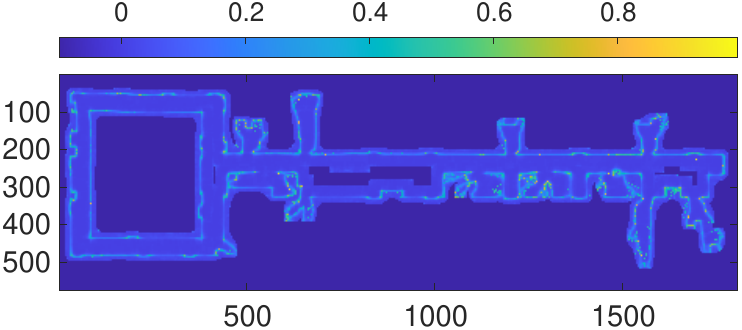}
        \caption{Heat map of $e$ for Map A produced from the autoencoder.}
        \label{fig:baseline-1-reconstruction}
    \end{subfigure}
    \begin{subfigure}{\columnwidth}
        \centering
        \includegraphics[width=\columnwidth]{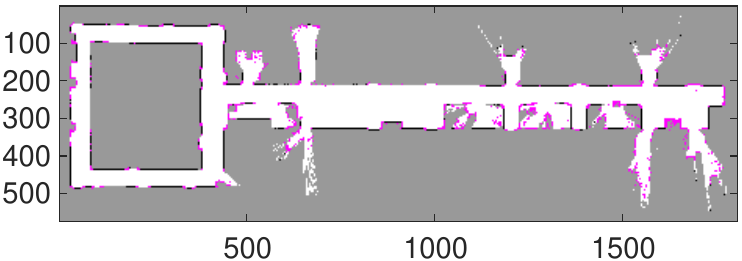}
        \caption{Grid map with cells classified as \textit{bad} highlighted.}
        \label{fig:baseline-1-threshold}
    \end{subfigure}
    
    \caption{Seattle grid map, a high quality occupancy grid map (Pre-2014 Robotics 2D-Laser Datasets (http://www.ipb.uni-bonn.de/datasets/): Seattle UW (D. Haehnel)) used in Andersone~\cite{andersone-2020-quality} and output from our method.}
    \label{fig:baseline-1-maps}
\end{figure}

We apply the developed method to two occupancy grid maps given in \cref{fig:baseline-1-maps,fig:baseline-2-occ-grid}. These two maps were also used in the map quality assessment method of Andersone~\cite{andersone-2020-quality} and we qualitatively compare our method to theirs as a baseline. To the best of our knowledge, Andersone is the most directly comparable method for reference-free map quality assessment.

\begin{figure}[htpb]
    \begin{subfigure}{\columnwidth}
        \centering
        \includegraphics[width=0.7\textwidth]{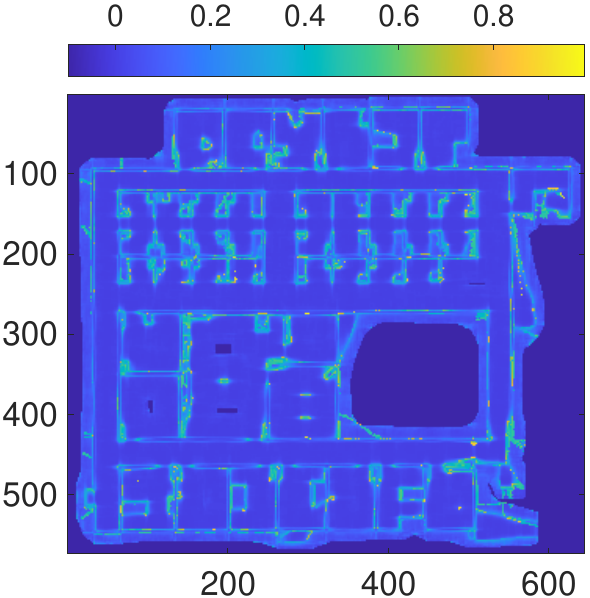}
        \caption{Heat map of $e$ produced from the autoencoder.}
        \label{fig:baseline-2-reconstruction}
    \end{subfigure}
    
    \begin{subfigure}{\columnwidth}
        \centering    
        \includegraphics[width=0.7\textwidth]{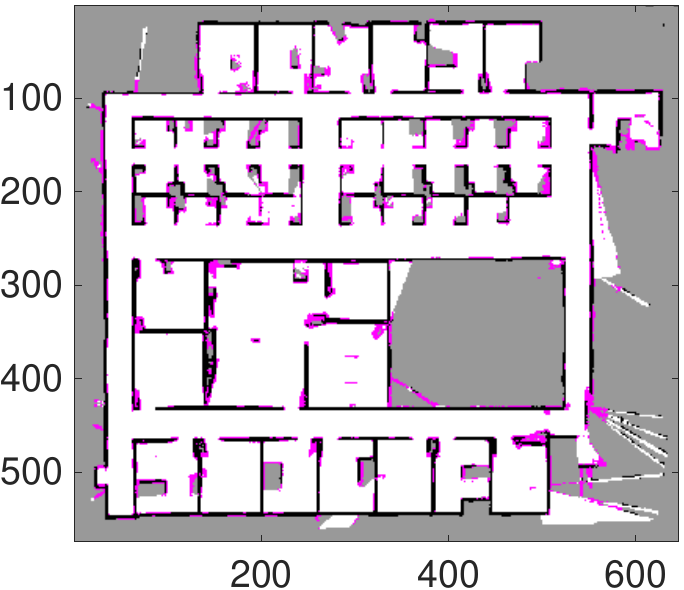}
        \caption{Grid map with cells classified as \textit{bad} highlighted.}        
        \label{fig:baseline-2-threshold}
    \end{subfigure}
    
    \caption{SDR site B grid map with cells classified as \textit{bad} highlighted.%
    }
    \label{fig:baseline-2-maps}
\end{figure}

In \cref{fig:baseline-2-maps} additional areas are classified as \textit{bad} compared to Andersone's. This difference could be due to our method being less restrictive when including a patch where only two cells or 0.2\% of cells for $32 \times 32$ patch size in the patch are \textit{occupied}. Andersone requires a patch to have 2.5\% occupied cell rate. This highlights our approach where insights are generated using patches which covers more of the map.
\begin{figure}%
    \centering
    \includegraphics[width=0.4\columnwidth]{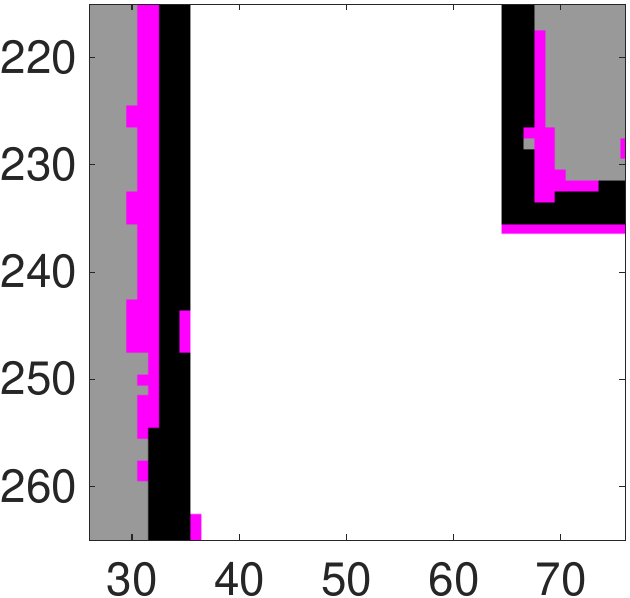}~
    \includegraphics[width=0.4\columnwidth]{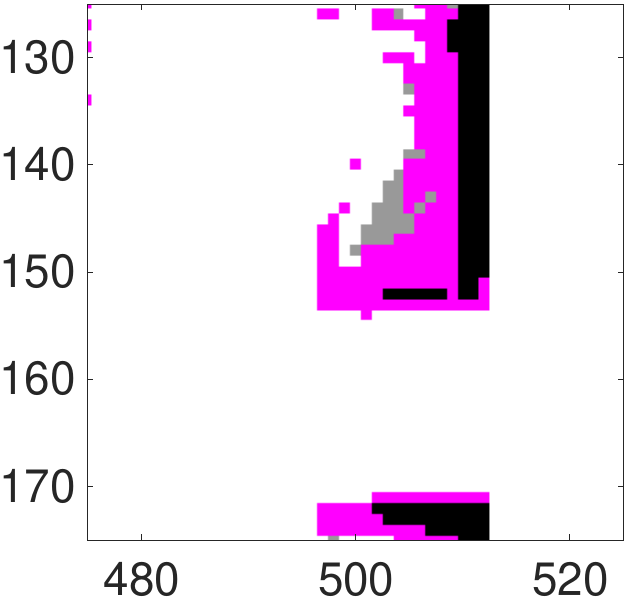} 
    \includegraphics[width=0.4\columnwidth]{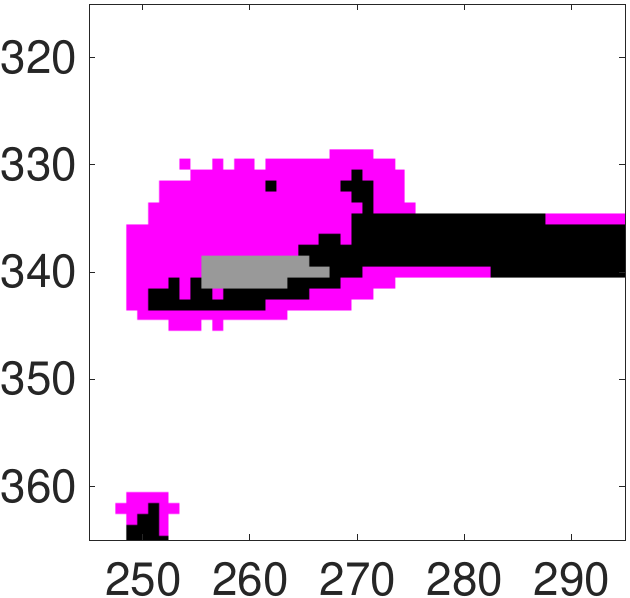}~
    \includegraphics[width=0.4\columnwidth]{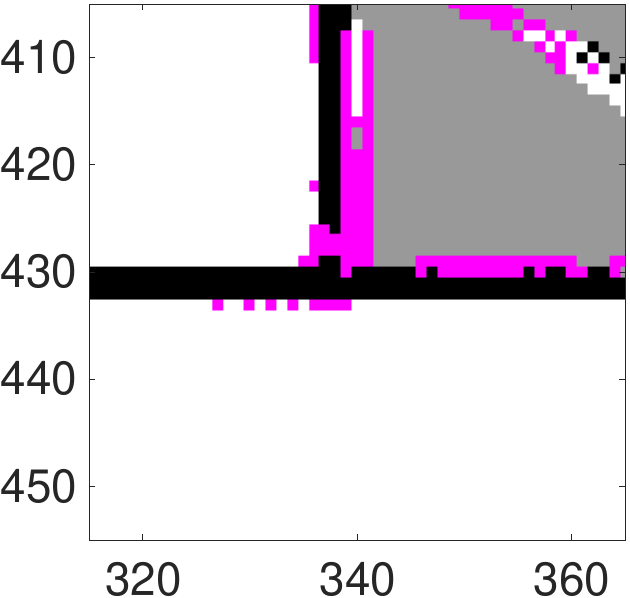}        
    \caption{Zoomed sections of SDR site B map (\cref{fig:Map-B-zoomed-regions}) with \textit{bad} cells highlighted.}
        \label{fig:baseline-2-zoomed-2}
\end{figure}

The model is trained with 53772 and 39032 patches for the SDR site B and Seattle maps, respectively. Training and inference times for the SDR site B map is 6.1 minutes and 10 seconds and for Seattle map is 4.3 minutes and 8 seconds.

Qualitatively, the \textit{bad} regions highlighted in \cref{fig:baseline-1-reconstruction,fig:baseline-2-reconstruction} are consistent with Andersone~\cite{andersone-2020-quality} but in contrast our method does not require manually labelled data.

\subsection{Evaluation in Other Environments}
\label{sec:warehouse-results}
\begin{figure}%
    \centering
    \includegraphics[width=0.85\columnwidth]{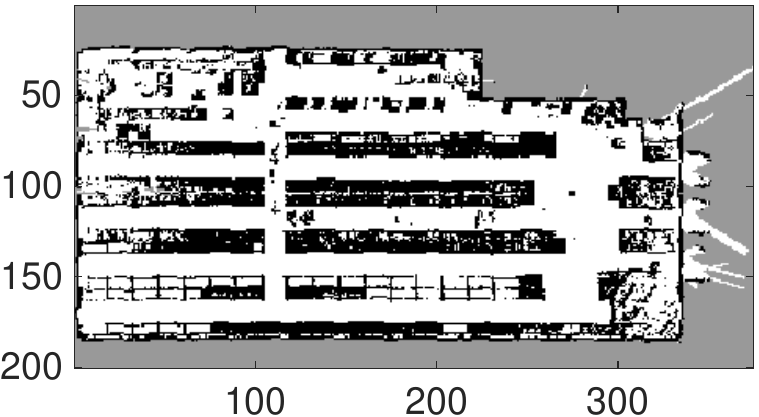}

    \includegraphics[width=0.85\columnwidth]{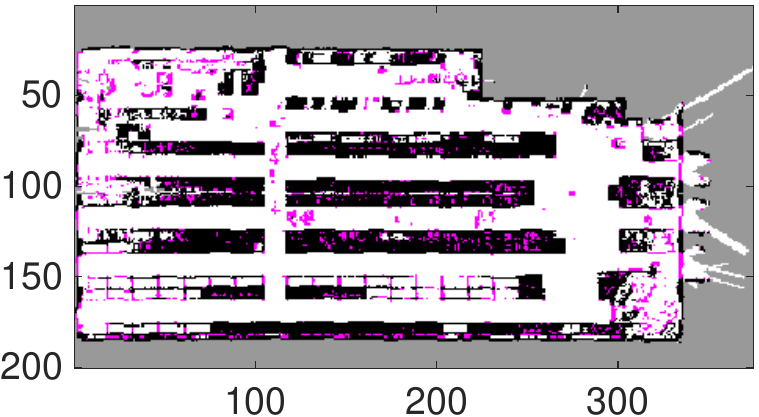}
    \caption{Baseline warehouse grid map (Top) and grid map with \textit{bad} cells highlighted (Bottom). A pillar is visible at map location $X=280$ and $Y=105$.}
    \label{fig:orkla-ground-truth}
\end{figure}

While the results from our method are promising for the environments in \cref{fig:baseline-1-occ-grid,fig:baseline-2-occ-grid} where we could make direct comparisons to a relevant learning-based baseline, we wanted to evaluate how the method performs in other types of environments where professional service robots are often employed, such as a warehouse for intralogistics or agile production applications. In a warehouse environment, it could be envisioned that the environment changes over time and is mapped frequently. In this scenario we evaluate maps to a model generated from a map that is considered to represent the type of environment well but is not necessarily entirely correct. This baseline warehouse map is shown in \cref{fig:orkla-ground-truth}.

Using the same approach demonstrated in \cref{sec:baseline-results}, map cells are classified as \textit{bad} using the cutoff threshold $e_{\tau}$ and can be observed in \cref{fig:orkla-ground-truth}. There are many regions highlighted and this should not be a cause for concern as this is expected. It is expected because also for mostly good maps, some areas have low coverage and map artifacts created by dynamic obstacles or clutter are present. The baseline warehouse model is trained using 47387 patches, training time is 4.8 minutes and inference time is 10 seconds.

Qualitatively assessing \cref{fig:orkla-ground-truth}, most of the \textit{bad} regions are appropriately labelled. The highlighted areas generally correspond to places with less sensor coverage where obstacles are less defined (near the left and lower-right parts of the map), clutter from moving obstacles that remains in the map in some of the aisles (in the central part of the map), intermittent features like the loading bays that are sometimes open and sometimes closed (on the right wall) and scattered points of free space inside obstacles where the Lidar sometimes see through and sometimes does not (e.g., within the large mostly black blocks of shelving in the central area of the map).

In future work we plan to close the loop by incorporating this map labeling such that
the resulting map of incorrect cells could be used as for planning a new mapping plan to validate the map. This would aid in correcting the misclassified cells in an automated fashion.

\begin{figure}%
    \centering
    \begin{subfigure}{\columnwidth}
        \includegraphics[width=0.465\textwidth]{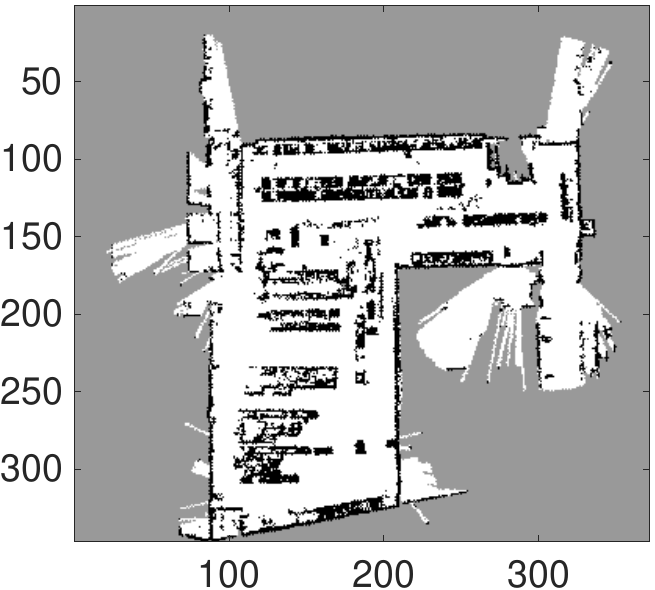} 
        \includegraphics[width=0.465\textwidth]{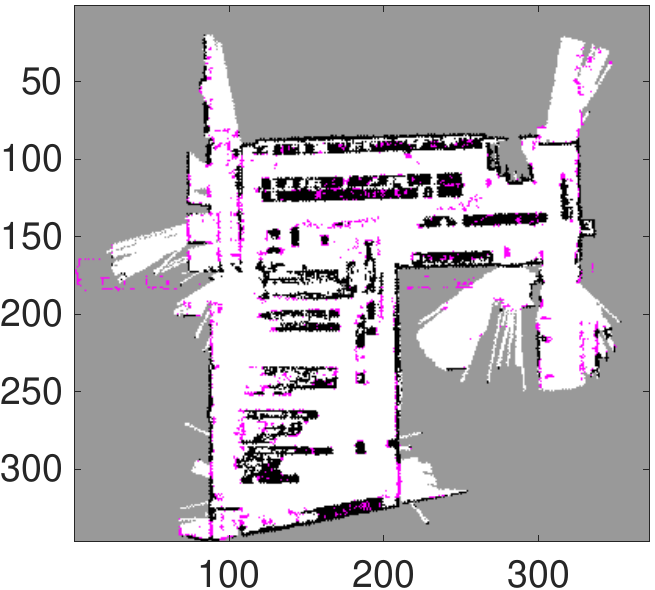}        
        \caption{Warehouse Map A grid map (left) and resulting \textit{bad} cells highlighted using the baseline warehouse map model (right).}
        \label{fig:orkla-1-occ-grid}
    \end{subfigure}
    \centering
    \begin{subfigure}{\columnwidth}
        \includegraphics[width=0.465\textwidth]{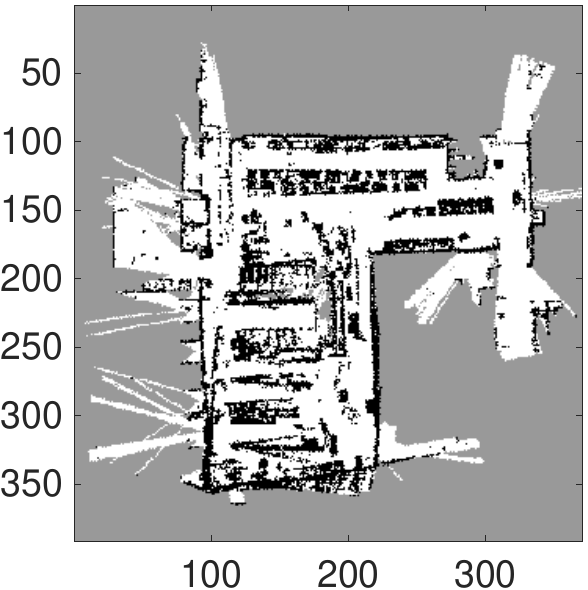}
        \includegraphics[width=0.465\textwidth]{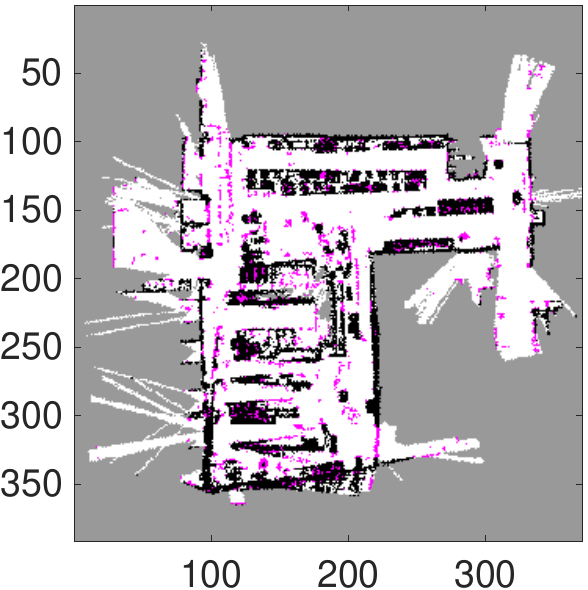}        
        \caption{Warehouse Map B grid map (left) and resulting \textit{bad} cells highlighted using the baseline warehouse map model (right).}
        \label{fig:orkla-2-occ-grid}
    \end{subfigure}
    \caption{Two warehouse maps evaluated using the model produced by the baseline warehouse map in \cref{fig:orkla-ground-truth}. The two maps are from a different warehouse from the baseline map.}
    \label{fig:orkla-extra-maps-occgrid-threshold}
\end{figure}

Two additional warehouse maps, labelled as Map A and Map B with different quality, are given in \cref{fig:orkla-1-occ-grid,fig:orkla-2-occ-grid}, respectively. These two maps are evaluated with an autoencoder model generated using the baseline warehouse map in \cref{fig:orkla-ground-truth}. Map B contains significant error due to poor localization in the lower half of the map. 

Training time is 4.5 minutes and inference time for each map is 10 seconds, highlighting the ability of our method to quickly provide a map quality estimate from a model.

\section{Discussion and Future Work}

The cut-off threshold $e_{\tau}$ is selected at the 95\% interval as described in \cref{sec:cell-classification} and given our hypothesis that majority of the map is accurate 95\% is reasonable. However, as it is a tunable parameter, one could imagine choosing the value with some prior knowledge of the mapping system.
Analyzing $e_{\tau}$ for the four different maps, the threshold is around 0.31--0.45 indicating the distribution of map cells appear to be similar.

The size of map patches $P$ should take into account the resolution of the overall map. Too small  map patches would introduce uninformative patches and too large patches may not generate a sufficient number of patches to learn a model or may not be locally consistent. Prior knowledge of the environment, if available, may be used to determine the patch size by estimating the expected physical size of structures to be observed and the grid map resolution.

Similar to image processing techniques, multiple models could be generated using different patch sizes with the results combined to possibly capture the low level structure in the map but also larger structures in the map. For example, identifying walls that are slightly warped may be possible with this approach and leverages still the method proposed in this paper.

A practical use case which, is left as future work for a map quality assessment tool could be for a ``self-commissioning" robot to autonomously identify regions of the map that need a second look. Actionable insights that could be generated after a mapping round and map evaluation could be for the robot to revisit the identified inaccurate areas for verification. This could aid in removing error due to clutter, sparsely mapped areas are explored and completed, and misclassified regions are corrected. For larger errors (due to bad loop closure) if the robot identifies errors are still in error, something more drastic may need to be done, such as to discard that region and redo the mapping for that area.

\section{Summary and Conclusion}

The process of editing maps places a large cognitive load on the editor and tools that can reduce the load is sought after in the robotics community and in industrial settings which utilize mobile robotic platforms. In these settings, significant resources are used to verify the map is good for repeatable localisation and this tool should help with that. Our data driven approach reformulates map quality assessment as anomaly detection since we exploit the majority of a map is correct or the map is reasonable. With this in mind, an autoencoder is able to represent the dominant structures in the map. Cell classification of \textit{good} or \textit{bad} used as a threshold the metric mean absolute reconstruction error that is estimated by comparing the input patch and its reconstruction using the model.

In our evaluation, the proposed approach identified the same \textit{bad} map cells compared to a baseline map quality assessment tool that required labelled map patches to train a classifier. The primary advantage of our method is that it requires \emph{no labels or ground truth} which can be difficult or infeasible to obtain. Additionally, the method was successful when assessing maps with various sources of error from environments not encountered during training. 

\bibliographystyle{IEEEtran}
\bibliography{papers}
\end{document}